\documentclass{SCIS2025}
\usepackage{threeparttable}
\usepackage{multirow}
\usepackage{color}
\usepackage{colortbl}
\usepackage{xcolor} 
\usepackage{graphicx}
\usepackage{amsmath,amssymb}
\usepackage{natbib}    
\usepackage{enumitem}

\begin{document}
\ArticleType{RESEARCH PAPER}
\Year{2025}
\Month{}
\Vol{68}
\No{}
\DOI{}
\ArtNo{000000}
\ReceiveDate{}
\ReviseDate{}
\AcceptDate{}
\OnlineDate{}
\AuthorMark{Ren W Q}
\AuthorCitation{Ren W Q, Wang W J, Zheng M, et al}



\title{In defense of the two-stage framework for open-set domain adaptive semantic segmentation}{In defense of the two-stage framework for open-set domain adaptive semantic segmentation}

\author[1]{Wenqi Ren}{}
\author[2]{Weijie Wang}{}
\author[3]{Meng Zheng}{}
\author[3]{Ziyan Wu}{}
\author[4]{Yang Tang}{yangtang@ecust.edu.cn} 
\author[5]{\\Zhun Zhong}{}
\author[2]{Nicu Sebe}{}


\address[1]{Shanghai Xiaoyuan Innovation Center, Shanghai 201108, China}
\address[2]{University of Trento, Trento 38100, Italy}
\address[3]{United Imaging Intelligence, Burlington 80807, United
States}
\address[4]{Key Laboratory of Advanced Control and
Optimization for Chemical Processes, \\ East China
University of Science and Technology, Shanghai 200237, China}
\address[5]{School of Computer Science and Information Engineering, Hefei University of Technology, Hefei 230000, China}

\abstract{Open-Set Domain Adaptation for Semantic Segmentation (OSDA-SS) presents a significant challenge, as it requires both domain adaptation for known classes and the distinction of unknowns. Existing methods attempt to address both tasks within a single unified stage. We question this design, as the annotation imbalance between known and unknown classes often leads to negative transfer of known classes and underfitting for unknowns. To overcome these issues, we propose \textbf{SATS}, a \textbf{S}eparating-then-\textbf{A}dapting \textbf{T}raining \textbf{S}trategy, which addresses OSDA-SS through two sequential steps: known/unknown separation and unknown-aware domain adaptation. By providing the model with more accurate and well-aligned unknown classes, our method ensures a balanced learning of discriminative features for both known and unknown classes, steering the model toward discovering truly unknown objects. Additionally, we present hard unknown exploration, an innovative data augmentation method that exposes the model to more challenging unknowns, strengthening its ability to capture more comprehensive understanding of target unknowns. We evaluate our method on public OSDA-SS benchmarks. Experimental results demonstrate that our method achieves a substantial advancement, with a +3.85\% H-Score improvement for GTA5$\rightarrow$Cityscapes and  +18.64\% for SYNTHIA$\rightarrow$Cityscapes, outperforming previous state-of-the-art methods.}


\maketitle

\section{Introduction} \label{sec:intro}

\begin{figure*}
	\centering
	\includegraphics[width=0.7\linewidth]{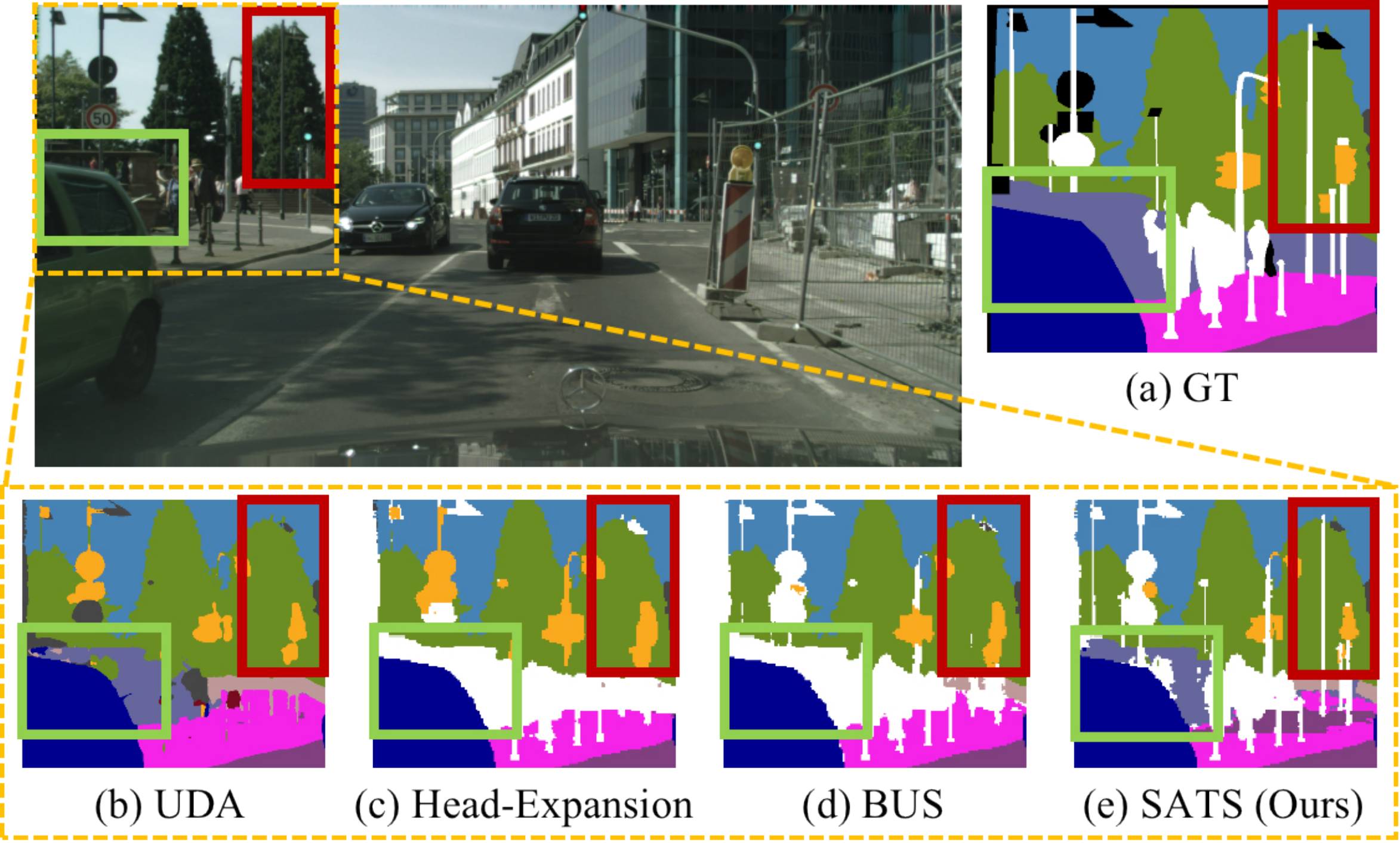}
	\caption{Visual comparison of the UDA method (MIC \cite{hoyer2023mic}), OSDA-SS baselines (head-expansion baseline and BUS \cite{choe2024open}), and our SATS under the OSDA-SS scenario. White pixels represent unknown classes. The UDA method (b) misclassifies all unknowns as known. Existing one-stage OSDA-SS approaches often relabel low-confidence pseudo-labeled pixels as unknown, leading to known classes being misclassified as unknown (highlighted in \textcolor[RGB]{146,208,80}{green boxes} of (c) and (d)). Additionally, because known classes are learned faster, they tend to overshadow unknown classes, leading to underfitting of these unknown classes (emphasized in \textcolor[RGB]{192,0,0}{red boxes} of (c) and (d)). Instead, our two-stage method (e) overcomes these issues, yielding more accurate segmentation.}
	\label{fig:motivation}
 \vspace{-10pt}
\end{figure*}

Recent advancements in semantic segmentation have achieved state-of-the-art (SOTA) performance \cite{chen2017deeplab, liu2019auto, xie2021segformer, strudel2021segmenter}. However, this success relies heavily on large labeled datasets, which require intensive annotation efforts \cite{cordts2016cityscapes, sakaridis2021acdc}. 
There have been continuous efforts \cite{ros2016synthia, richter2016playing} leveraging synthetic datasets with automatically generated annotations to alleviate this issue.
Yet, due to domain gaps between synthetic and real-world data, models trained on synthetic datasets (source domain) often experience compromised performance in real-world scenarios (target domain). Unsupervised Domain Adaptation (UDA) \cite{tranheden2021dacs, hoyer2022daformer, hoyer2022hrda, hoyer2023mic} has been proposed to bridge domain gaps for semantic segmentation, enabling models trained on labeled source domains to generalize to unlabeled target domains. Nonetheless, most UDA methods typically assume a \emph{closed-set} setting, where the source and target domains share the same set of classes ($C_S = C_T$). 
This assumption becomes violated when the target domain introduces new classes, leaving these methods unable to classify the novel classes (see Figure \ref{fig:motivation}(b)). This issue leads us to investigate \emph{Open-Set Domain Adaptation in Semantic Segmentation (OSDA-SS)}, where the target domain includes novel, private classes unseen in the source domain ($C_S \subset C_T$). 
OSDA-SS is even challenging as it requires solutions to handle both domain adaptation for known classes and the separation of unknowns, \emph{i.e.}, effectively identify each known class while assigning a single \emph{unknown} label to any target-specific private classes.

To tackle OSDA-SS, a solid baseline can be developed by extending conventional UDA methods \cite{saito2018open, saito2020universal, choe2024open}, with an additional dimension in the classifier head for unknown classes and reassignment of low-confidence pixels as unknown during pseudo-labeling. 
While effective, this head-expansion pipeline still has limitations that hamper the performance of OSDA-SS. \textbf{(1) Negative transfer of known classes.} Without annotations for unknowns, the expanded head fails to establish clear boundaries, especially at the beginning of the training. This causes ambiguous or low-confidence regions—whether known or unknown—being classified as unknowns (see \textcolor[RGB]{146,208,80}{green boxes} of Figure \ref{fig:motivation}(c)), resulting in negative transfer of known classes. \textbf{(2) Underfitting of unknown classes.} The annotation imbalance, with well-annotated known classes but noisy pseudo labels for unknowns, causes models to prioritize learning of known classes~\cite{liu2020negative, cao2022openworld}. This naturally leads to higher accuracy on known classes but underfitting of unknowns, causing misclassification between unknown/known classes (see \textcolor[RGB]{192,0,0}{red boxes} in Figure \ref{fig:motivation}(c)). Though attempts have been made in methods like BUS \cite{choe2024open} to alleviate the issue by data mixing, limitations remain due to the inherent noise in pseudo labels (see Figure \ref{fig:motivation}(d)). 

In this paper, we argue that the above limitations stem from \emph{jointly implementing unknown separation and domain adaptation within a single stage}. To this end, we propose a Separating-then-Adapting Training Strategy (SATS) for solving the above limitations, which divides the OSDA-SS problem into two sequential stages, unknown detection and domain adaptation. \textit{In the first stage}, we focus on developing an effective expanded head for identifying unknowns. 
To provide sufficient, correct supervision for unknown classes, we propose explicitly constructing ``virtual unknowns'' (known unknowns) within source samples,  instead of relying on noisy pseudo labels as in the previous method \cite{choe2024open}. By generating irregular, arbitrary-colored regions that mimic target unknowns, we enable effective learning of the expanded head and ensure more robust generalization to target unknowns. Unlike the first stage, \textit{the second stage} emphasizes domain adaptation with both known and unknown classes. This process starts with pre-training under a self-training framework using both source and target domains, in which the source data is additionally augmented by the high-confidence unknowns identified by the unknown detection model of the first stage. In this way, we enable the model to not only treat known and unknown classes more equally but also optimize with more accurate unknown samples during training, enhancing its ability to differentiate truly unknown classes. Following the pre-training, we extend the framework to further explore ``hard unknowns'' that are easily overwhelmed by known classes. By identifying these challenging unknowns and using them to dynamically refine source unknowns, the model is further improved to reject complex unknowns. We evaluate our method on two synthetic-to-real OSDA benchmarks. The results demonstrate significant improvements over previous approaches (see Figure \ref{fig:motivation}(e)). For instance, our method achieves a +5.51\% IoU gain for unknowns on GTA5 $\rightarrow$ Cityscapes and another +26.37\% IoU on SYNTHIA $\rightarrow$ Cityscapes.  In summary, our key contributions are as follows:
\begin{itemize}
\item We propose a Separating-then-Adapting Training Strategy (SATS) for OSDA-SS, effectively minimizing negative transfer of known classes and reducing underfitting for unknowns.
\item We propose the construction of virtual unknowns and the exploration of hard unknowns, helping the model achieve robust generalization to target unknowns.
\item Our proposed method significantly outperforms the previous approaches, setting a new SOTA performance on OSDA-SS benchmarks. Moreover, our SATS can be seamlessly embedded into previous one-stage methods, leading to consistent improvements.
\end{itemize}

\section{Related work}
\label{sec:relatedwork}

\subsection{Closed-set domain adaptation} 
Given a shared class space ($C_S=C_T$), closed-set domain adaptation (CSDA) seeks to adapt a semantic segmentation model trained on a labeled source domain to an unlabeled target domain, with adversarial training and self-training being the primary approaches. The first group adopts a learnable domain discriminator to offer supervision within a GAN framework \cite{goodfellow2014generative}, aiming to reduce domain discrepancies in inputs \cite{hoffman2018cycada, pizzati2020domain, gong2021dlow}, features \cite{ganin2016domain, long2018conditional, tsai2018learning}, outputs \cite{saito2018maximum, vu2019advent, luo2021category}, or patches \cite{tsai2019domain}. In self-training, high-confidence pseudo labels are predicted based on confidence thresholds \cite{zhang2018collaborative, zou2018unsupervised, mei2020instance} or class prototypes \cite{zhang2021prototypical, zhang2019category} for the target domain. To stabilize training, consistency regularization \cite{araslanov2021self, choi2019self} is frequently used across different data augmentations \cite{melas2021pixmatch, araslanov2021self, choi2019self}, domain mixup \cite{tranheden2021dacs, zhou2022context, hoyer2022daformer, kim2023bidirectional}, varying context \cite{zhou2022context, hoyer2022hrda}, or multiple models \cite{zhang2021multiple, zhou2022uncertainty, zheng2021rectifying}. Several studies also tackle the CSDA challenge by combining adversarial training with self-training \cite{wang2020classes, wang2020classes, kim2020learning, zheng2021unsupervised}, refining boundaries \cite{liu2021bapa}, or reducing domain gaps through contrastive learning \cite{huang2022category, xie2023sepico}. Despite these advancements, CSDA methods face limitations in real-world applications due to the assumption of a shared class space. This constraint becomes particularly problematic when the target domain includes unknown classes, leading to frequent performance drops \cite{kundu2020towards}. This highlights the need for techniques that can handle both known and unknown classes, offering more flexibility and robustness in real-world scenarios.

\subsection{Open-set domain adaptation} 
Open-set domain adaptation (OSDA) represents a more practical variation of CSDA, where the target domain is allowed to contain a set of new private classes that do not exist in the source domain ($C_S\subset C_T$) \cite{panareda2017open}. The objective of OSDA is to accurately classify the known classes present in the source domain while identifying \emph{any} new classes unique to the target domain as ``\emph{unknown}'' \cite{saito2018open}. To date, most OSDA research has concentrated on classification tasks
\cite{luo2020progressive, liu2019separate, luo2020progressive, wang2024progressively}. However, OSDA has received limited attention in semantic segmentation tasks. To the best of our knowledge, BUS \cite{choe2024open} is the only notable effort focused on this area. This approach develops a head-expansion framework--adding an extra dimension to the classifier head to isolate unknown classes and reclassifying low-confidence pixels as unknown during pseudo-label generation--illustrating effectiveness in rejecting unknowns. Nevertheless, this one-stage method combines the separation of unknowns and domain adaptation for known classes within a single stage, resulting in the negative transfer of known classes and underfitting of unknowns. Therefore, effectively tackling OSDA-SS remains an unresolved  and critical challenge in the field.

\section{Method}

\begin{figure*}
	\centering
	\includegraphics[width=\linewidth]{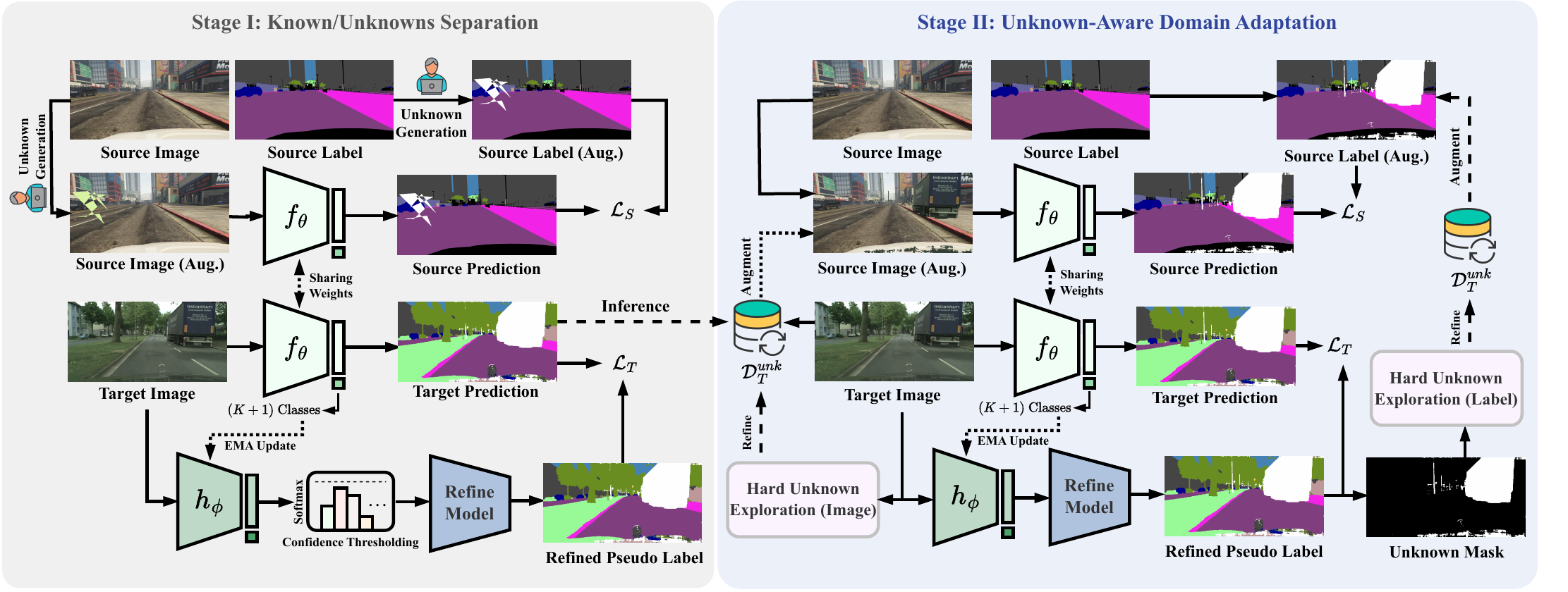}
	\caption{\textbf{Illustration of our proposed SATS method}, which comprises two sequential stages: known/unknown separation and unknown-aware domain adaptation. \textbf{Known/unknown separation} (Section \ref{sec:stage1}) aims to learn an expanded head to accurately identify unknown classes. To this end, ``virtual unknowns'' are constructed within source samples, providing reliable supervision for these unknown classes. \textbf{Unknown-aware domain adaptation} (Section \ref{sec:stage2}) begins with pre-training on both source and target domains, where the source data is further enriched with high-confidence unknowns identified from the first stage. This approach balances the learning of known and unknown classes, allowing the pipeline to further explore ``hard unknowns'' for improved robustness.}
	\label{fig:framework}
 \vspace{-5pt}
\end{figure*}

\subsection{Task statement}
In OSDA-SS, we have access to labeled source data, denoted as $\mathcal{D_S}=\{(x_s^n,y_s^n)\}_{n=1}^{N_s}$, and unlabeled target data, denoted as $\mathcal{D_T}=\{x_t^n\}_{n=1}^{N_t}$, where the source and target domains are drawn from distinct distributions ($P_S\neq P_T$).  Here, $x$ represents an RGB image, and $y$ denotes the corresponding pixel-wise semantic label. The source and target domains share a common set of $K$ known classes $C_S$. Additionally, the target domain also contains an additional set of $K'$ private novel classes $C_{T\setminus S}$, which are not present in the source domain and should be uniformly considered as ``\emph{unknown}'' (class $K+1$) \cite{panareda2017open,saito2018open,choe2024open}. Therefore, the goal of OSDA-SS is to train a segmentation model $f_\theta$ on both $\mathcal{D_S}$ and $\mathcal{D_T}$, with the expectation that the trained model can segment either one of the known classes or the unknown class in the target domain. This involves addressing two key challenges: 1) the separation of unknown classes ($C_{T\setminus S}\neq \emptyset$) and 2) the domain adaptation within known classes ($P_S\neq P_T$ within $C_S$).

\subsection{Framework overview}
In this paper, we propose SATS, a Separating-then-Adapting Training Strategy for addressing OSDA-SS through two sequential steps: known/unknown separation and unknown-aware domain adaptation. 
As shown in Figure \ref{fig:framework}, in the first stage, we train a $(K+1)$-class classifier head to separate target samples into known and unknown classes. To achieve this, we generate ``virtual unknowns'' within source samples, providing the $(K+1)$-class classifier head with sufficient and accurate supervision to effectively handle unknowns (Section \ref{sec:stage1}). The second stage focuses on domain adaptation for both known and unknown classes, which is based on a self-training framework. This process begins with pre-training on both source and target domains, with the source data further enhanced by high-confidence unknowns identified by the unknown detection model from the first stage. This approach provides the model with more accurate unknown samples, enabling a balanced representation of both known and unknown classes, which improves its ability to distinguish truly unknown classes. Following the pre-training, we extend the pipeline to further explore ``hard unknowns'' that are easily overwhelmed by known classes. By identifying these challenging unknowns and using them to dynamically augment source unknowns, the model is further improved to reject complex unknowns (Section \ref{sec:stage2}). In the following, we present the detailed description of our two stages, separately.

\subsection{Stage I: known/unknown separation}
\label{sec:stage1}
To effectively address the OSDA-SS problem, the first stage focuses on the essential task of distinguishing between known and unknown classes. This separation is crucial for enabling targeted adaptation strategies, as it can expose the model to more accurate and well-aligned unknown samples in the second stage. However, the lack of labels for unknown classes causes negative transfer, resulting in many known classes being misclassified as unknowns. Using predicted unknowns in noisy pseudo labels has been considered as a potential solution but can further exacerbate this issue \cite{choe2024open}. To overcome these challenges, we propose creating ``virtual unknowns'' within source samples, providing precise supervision that enables the classifier to handle unknowns more effectively. The detailed process for implementing this approach is outlined below.

\noindent\textbf{Head-expansion baseline.} In this stage, we establish our baseline by extending the self-training CSDA framework into the head-expansion baseline, which has proven effective in isolating unknown classes \cite{choe2024open}. This extension involves expanding the classifier head from $K$ to $(K+1)$ classes and assigning low-confidence pixels to the unknown class when generating pseudo labels. Specifically, a neural network $f_\theta$ is trained on the labeled source domain using a supervised cross-entropy loss $\mathcal{L}_S$ as follows:
\begin{equation}
	\mathcal{L}_S = -\sum\limits_{i = 1}^H\sum\limits_{j = 1}^W\sum\limits_{k= 1}^{K+1} y_s^{(i,j,k)}\log f_\theta(x_s)^{(i,j,k)}.
	\label{eq:supervised_loss}
    \vspace{-5pt}
\end{equation} 
In Equation \ref{eq:supervised_loss}, $i$ and $j$ denote pixel coordinates within the image, while $k$ represents the class index. To bridge the domain gap between the source and target domains, an unsupervised loss $\mathcal{L}_T$ is formulated for the target samples, using a teacher network $h_{\phi}$ to generate pseudo labels $\hat{y}_t$:
\begin{equation}
	\mathcal{L}_T = -\sum\limits_{i = 1}^H\sum\limits_{j = 1}^W\sum\limits_{k= 1}^{K+1} q_t\hat{y}_t^{(i,j,k)}\log f_\theta(x_t)^{(i,j,k)}.
	\label{eq:unsupervised_loss}
\end{equation} 
The pseudo labels are generated using the following rule:
\begin{equation}
	\hat{y}_{t}^{(i,j)} = \begin{cases} 
k, & \text{if } \left( \max_{k\in C_S} h_{\phi}(x_t)^{(i,j,k)} \geq \tau_1 \right), \\
K + 1, & \text{otherwise}.
\end{cases}
	\label{eq:pseudo_label1}
\end{equation} 
Here, $\tau_1$ is a predefined threshold used to assign pixels to the ``\emph{unknown}'' class if their maximum softmax probability falls below $\tau_1$. $q_t$ is a confidence weighting factor that estimates the quality of the pseudo labels \cite{hoyer2022daformer,hoyer2022hrda,hoyer2023mic} with a predefined threshold $\tau_2$:
\begin{equation}
	q_t = \frac{1}{H \times W}\sum\limits_{i = 1}^H\sum\limits_{j = 1}^W \mathbb{I} \left[ \max_{k\in C_T} h_{\phi} (x_t )^{(i,j,k)} > \tau_2 \right].
	\label{eq:confidence}
\end{equation} 
We also use a frozen refinement model \cite{kirillov2023segment} to enhance the pseudo labels based on class-ratio statistics \cite{choe2024open}. To further stabilize the learning process, the weights of the teacher model $h_{\phi}$ are updated as the exponential moving average (EMA) of the weights of the student model $f_{\theta}$ after each training iteration. The EMA update rule is governed by a smoothing factor $\alpha$:
\begin{equation}
	\phi \gets \alpha * \phi+ (1-\alpha)*\theta.
	\label{eq:ema}
\end{equation} 

\noindent\textbf{Virtual unknown construction.} Building on the head-expansion baseline, we train the expanded head with target pseudo labels that include unknown labels. However, a significant challenge arises as the source domain does not contain any unknown classes. Thus, the expanded head cannot be updated with source data, leading to inefficiencies in the training process. BUS \cite{choe2024open} addresses this by using noisy pseudo labels, but it exacerbates negative transfer, causing some known classes to be misclassified as unknown.

In contrast to BUS, we propose an innovative strategy that enhances source data by generating ``virtual unknowns''. This approach introduces random, irregular shapes within source domain images $x_s$ and fills these regions with arbitrary colors, thereby creating ``virtual-unknown'' areas that mimic target unknown classes. Specifically, the implementation includes three steps: a) We randomly sample a set of pixel coordinates from a source image to define the vertices, and then connect them in order to form the polygon. Here, we ensure that the last vertex connects back to the first. b) We use the scanline algorithm to fill the polygon with a random color. c) The size of the polygon is adjusted by a scale factor $\gamma$ to balance the proportion of knowns and virtual unknowns in source domain.

Let us define a binary mask $m_a$ that spatially localizes the generated virtual unknown regions within the source image $x_s$, where a pixel value of 1 corresponds to the artificially constructed unknown regions and 0 otherwise. The integration of these virtual unknown regions into the source image is formally achieved through the following composition operation:
\begin{equation}
    \tilde{x}_s = x_s \odot (1 - m_a) + c \cdot m_a,
\end{equation}
\begin{equation}
    \tilde{y}_s = y_s \odot (1 - m_a) + (K+1) \cdot m_a.
\end{equation}
Here, $\tilde{x}_s$ and $\tilde{x}_s$ denote the augmented source image and its corresponding label. $c$ represents a randomly chosen color vector that simulates arbitrary textures or appearances. The symbol $\odot$ denotes element-wise multiplication. Note that although virtual unknowns cannot fully represent target unknowns, their inclusion enhances the model’s generalization ability. First, the diversity of virtual unknowns on shape and color helps prevent overfitting to specific features of known classes. Second, they allow the model to learn the boundary between known and unknown classes by treating virtual unknowns as ambiguous objects outside the known classes.
By doing this, we obtain an unknown detection model $f_{\theta}^\dag$ that can effectively identify unknowns. We utilize $f_{\theta}^{\dag}$ to infer target samples, resulting in a set of identified target-unknown classes $\mathcal{D}_\mathcal{T}^{unk} = \{(x_{t,unk}^n, \hat{y}_{t,unk}^n)\}_{n=1}^{N_t}$. Here, $x_{t,unk}$ and $\hat{y}_{t,unk}$ represent the input image and output segmentation maps of $f_{\theta}^{\dag}$.

\subsection{Stage II: unknown-aware domain adaptation}
\label{sec:stage2}
Upon completing Stage I, we effectively isolate the unknown classes from the known ones, setting the foundation for conducting the domain adaptation process. Here, we introduce this process as follows.

\noindent\textbf{UDA framework construction.} Our domain adaptation stage is based on the self-training CSDA framework, which can be transformed by adjusting the pseudo label generation (Eq. \ref{eq:pseudo_label1}) in the head-expansion baseline via:
\begin{equation}
    \hat{y}_t^{(i,j)} = \max_{k\in C_T} h_{\phi}(x_t)^{(i,j,k)}.
    \label{eq:pseudo_label2}
\end{equation}
To meet the assumption of the CSDA framework, we construct a closed-set scenario by augmenting the source data with high-quality unknowns identified in the first stage. This ensures that both the source and target domains share $(K+1)$ classes, allowing us to seamlessly apply self-training CSDA methods. Additionally, it increases the diversity of the training data, facilitating cross-domain adaptation for both known and unknown classes. Specifically, we first identify the regions associated with unknown classes in each target domain image $x_{t,unk}$, using the segmentation map $\hat{y}_{t,unk}$. In $\hat{y}_{t,unk}$, pixels with a value of $(K+1)$ indicate the unknown classes. To specifically isolate these pixels, we generate a binary mask, $m_{t,unk}$, which highlights only the target unknown class regions:
\begin{equation}
    m_{t,unk}^{(i,j)} = \begin{cases}
        1, & \text{if } \hat{y}_{t,unk}^{(i,j)} = K+1, \\
        0, & \text{otherwise}.
\end{cases}
\end{equation}
Using $m_{t,unk}$, we then perform a class mixup procedure \cite{tranheden2021dacs} that blends information from both the source image $x_s$ and the target image $x_{t,unk}$, as well as their corresponding labels $y_s$ and $\hat{y}_{t,unk}$. The augmented source image $\tilde{x}_s$ and its corresponding label $\tilde{y}_s$ are generated as follows:
\begin{equation}
    \tilde{x}_s = m_{t,unk} \odot x_{t,unk} + (1-m_{t,unk}) \odot x_s,
\end{equation}
\begin{equation}
    \tilde{y}_s = m_{t,unk} \cdot (K+1) + (1-m_{t,unk}) \odot y_s.
\end{equation}

\noindent\textbf{Pre-training and hard unknown exploration.} Using the self-training framework, we begin with pre-training on both the target domain and the reconstructed source domain. This approach allows balanced learning of known and unknown classes and optimizes training with more accurate unknown samples, enhancing the model’s ability to distinguish truly unknown classes. After pre-training, the model generates more discriminative features, enabling it to uncover additional unknowns in target samples. Motivated by this, we extend this framework to further investigate ``hard unknowns''--those that are easily overshadowed by known classes. We observe that these hard unknown classes are often misclassified as dominant/head known classes. Based on this observation, we propose utilizing the stage II pseudo label $\hat{y}_t$ to dynamically refine $m_{t,unk}$ so that we can provide the model with a more comprehensive distribution of target unknowns. Specifically, given a pseudo label $\hat{y}_t$ predicted by the stage II model, we update the unknown mask $m_{t,nuk}$ as follows:
\begin{equation}
    m_{t,nuk}^{(i,j)} =
        \begin{cases}
            1, & \text{if } \hat{y}_{t,unk}^{(i,j)} = K+1 \text{ or } 
             (\hat{y}_t^{(i,j)} = K+1 \text{ and } \hat{y}_{t,unk}^{(i,j)} \in C_H), \\
            0, & \text{otherwise}.
        \end{cases}
\end{equation} 
Here, $C_H$ represents the set of head known classes. In this way, we not only recover the appearance and shape of source unknowns but also increase the diversity of challenging unknown classes, improving the model’s ability to handle complex unknowns.

\begin{table*}[t]
	\begin{center}
		\fontsize{7}{7}\selectfont
			\caption{Comparison of results with various competing methods on two benchmarks. ``-C'' and ``-H'' denote the confidence-threshold baseline and the head-expansion baseline, respectively. The best results are \textbf{in bold}.}
		\label{tab:STOA0}
        \setlength{\tabcolsep}{0.8pt}
		\begin{tabular}{c|ccccccccccccc|>{\columncolor{gray!20}}c>{\columncolor{gray!20}}c>{\columncolor{gray!20}}c}
			\toprule
            \multicolumn{17}{c}{\textbf{GTA5 $\rightarrow$ Cityscapes}}\\ \midrule
			Method  & Road & S.walk & Build. & Wall & Fence & Light & Veget. & Terrain & Sky   & Car   & Bus    & M.bike & Bike   & Common & Private & H-Score\\ \midrule
        OSBP \cite{saito2018open}        & 4.92 & 3.93   & 42.80  & 2.55 & 6.04  & 14.29 & 68.58  & 26.50   & 44.21 & 41.78 & 0.94   & 7.20   & 3.42   & 20.55  & 4.49    & 7.34  \\
        UAN \cite{you2019universal}      &65.97 &23.41   & 76.41  &37.26 &18.50  &20.13  &80.57   &30.37    &82.47  &77.35  &27.80   &16.62   &0.00    &38.00   &3.59     &6.56\\
        UniOT \cite{jang2022unknown}     &17.67 &5.14    & 44.86  &55.45 &2.31   &52.61  &40.01   &3.37     &79.43  &52.87  &52.31   &7.18    &0.00    &20.20   &5.36     &7.49\\ \midrule
        ASN-C  \cite{tsai2018learning}      &82.34 &2.21    & 75.30  &8.01  &3.52   &9.99   &71.96   &15.61    &70.97  &77.16  &22.59   &20.80   &0.06    &35.43   &10.84    &16.60\\
        Pixmatch-C  \cite{melas2021pixmatch}&79.27 &2.06    & 72.36  &6.96  &2.94   &11.07  &76.29   &23.23    &77.72  &79.77  &44.72   &18.02   &0.01    &38.03   &9.46     &15.15\\
        DAF-C  \cite{hoyer2022daformer}     &94.26 &48.69   & 83.47  &38.67 &32.83  &41.71  &87.79   &39.15    &93.59  &85.29  &47.04   &28.36   &46.86   &61.26   &14.63    &23.36\\
        HRDA-C  \cite{hoyer2022hrda}        &95.14 &62.58   & 82.92  &47.44 &43.57  &53.18  &88.26   &44.42    &92.92  &90.23  &57.43   &14.71   &56.83   &63.82   &12.13    &20.39\\
        MIC-C  \cite{hoyer2023mic}          &93.26 &58.96   & 79.30  &21.62 &31.41  &39.32  &85.48   &31.94    &91.64  &88.16  &44.77   &47.64   &42.77   &58.17   &11.87    &19.71\\ 
        DAF-H  \cite{hoyer2022daformer}&95.80 &65.37   & 87.12  &54.08 &45.81  &51.78  &89.20   &42.93    &91.03  &89.19  &37.93   &50.54   &48.49   &66.09   &29.23    &40.53\\
        HRDA-H   \cite{hoyer2022hrda}  &95.31 &37.70   & 89.26  &57.41 &37.00  &\textbf{61.16}  &90.96   &46.86    &\textbf{94.39}  &93.39  &62.45   &58.13   &65.71   &68.44   &31.02    &42.70\\
        MIC-H   \cite{hoyer2023mic}    &\textbf{97.14} &\textbf{79.45}   & 88.78  &55.6  &53.92  &26.11  &89.94   &50.98    &93.54  &92.46  &\textbf{69.09}   &54.53   &63.43   &70.38   &31.78    &43.79\\ \midrule
        BUS \cite{choe2024open} (DAF)                        &91.90 &41.06   & 88.04  &48.65 &48.74  &48.94  &89.59   &44.37    &91.61  &89.99  &46.09   &48.49   &62.47   &64.61   &39.23    &48.82\\
        BUS \cite{choe2024open} (HRDA)                       &88.07 &39.59   & 88.57  &55.12 &48.29  &56.24  &90.02   &46.30    &91.76  &92.03  &46.96   &57.10   &66.02   &66.62   &42.50    &51.89\\
        BUS \cite{choe2024open} (MIC)                        &95.06 &66.65   & 90.53  &55.37 &55.38  &57.20  &91.12   &49.69    &92.96  &93.50  &68.81   &58.73   &67.04   &72.47   &55.42    &62.81\\
        Ours (DAF)                       &94.45 & 59.80 & 88.57 & 50.49 & 46.67 & 51.26 & 89.59 & 46.80 & 91.42 & 90.89 & 42.68 & 52.74 & 65.41   &66.98 &50.32& 57.47\\
        Ours (HRDA)                      &95.99 & 71.23 & 89.59 & 60.67 & 43.62 & 57.06 & 90.31 & 50.86 & 92.82 & 91.39 & 42.06 & 51.29 & 70.51   &69.80   &55.99    &62.14\\
        Ours (MIC)                       &96.14 & 75.30 & \textbf{90.82} & \textbf{61.27} & \textbf{57.12} & 60.90 & \textbf{91.60} & \textbf{54.49} & 93.68 & \textbf{93.72} & 45.88 & \textbf{62.41} & \textbf{73.03}   &\textbf{73.57}   &\textbf{60.93}    &\textbf{66.66}\\ 
			\bottomrule
		\end{tabular}

        \vspace{3pt}
        \setlength{\tabcolsep}{1.7pt}
        \begin{tabular}{c|cccccccccccc|>{\columncolor{gray!20}}c>{\columncolor{gray!20}}c>{\columncolor{gray!20}}c}
			\toprule
            \multicolumn{16}{c}{\textbf{SYNTHIA $\rightarrow$ Cityscapes}}\\ \midrule
			Method  & Road & S.walk & Build. & Wall & Fence & Light & Veget.  & Sky   & Car   & Bus    & M.bike & Bike   & Common & Private & H-Score\\ \midrule
        OSBP \cite{saito2018open}        & 6.71 & 9.49   & 49.83  & 0.70 & 0.00  & 0.76 & 26.03    & 36.91    & 20.04  &  4.76  &  2.90   &  8.70   & 13.20   & 4.90  & 7.14 \\
        UAN \cite{you2019universal}      &33.24 &19.03   & 71.49  & 4.02 & 0.05  & 14.34  &75.78   & 81.06    & 53.88  & 19.34  & 8.14    &21.84    &31.30    &4.53   &7.91 \\
        UniOT \cite{jang2022unknown}     &0.00  &16.79   & 18.52  & 1.05 &6.49   & 16.80  & 14.52   & 57.40   & 6.48   & 2.59   & 3.73    & 3.88    &12.35    &5.49   &7.06 \\ \midrule
        ASN-C  \cite{tsai2018learning}      &72.70 &41.29   &  73.59  &7.38  & 0.08   & 1.17   &71.35   & 82.22    &67.35  &23.30  & 0.94   &20.56    &38.49    &4.62   &8.25 \\
        Pixmatch-C  \cite{melas2021pixmatch}&74.16 &8.15    &  76.21  & 0.01  & 0.00  & 5.64  & 44.15   & 63.76    & 44.66  & 17.27  & 0.13   &0.38   &26.30    &6.87   &11.00 \\
        DAF-C  \cite{hoyer2022daformer}     &70.10 &39.65   &  83.09  &22.75 &4.66  &41.19  & 81.56   & 91.79    & 84.36  & 51.13  &43.78   &46.20   &51.49   &9.07   &15.57\\
        HRDA-C  \cite{hoyer2022hrda}        &85.62 &41.74   &  83.29  & 36.35 & 0.86  &35.17  & 83.98   & 90.90    & 84.74  &50.42  & 46.78   & 58.33   &54.68   &12.68   &20.82\\
        MIC-C  \cite{hoyer2023mic}          &88.31 &\textbf{70.71}   & 85.00  & 26.23 & 6.60  &35.27  & 84.80   & 91.41    & 81.47  &53.62  & 55.39   & 58.20   &57.46   &10.02   &17.23\\ 
        DAF-H  \cite{hoyer2022daformer}&82.93 &49.26   & 86.71  &39.21 &7.15  &52.35  &77.15   &88.26    &87.02  &63.00 &54.37   &52.84   &61.69   &32.75   &42.79   \\
        HRDA-H   \cite{hoyer2022hrda}  &87.13 & 35.31 & 86.22 & 41.08 & 5.12 & 40.27 & 86.30 & 92.59 & 89.64 & 66.93 & 57.30 & 59.09   &62.25   &23.74   &36.40  \\
        MIC-H   \cite{hoyer2023mic}    &\textbf{89.08} & 58.55 & 86.01 & 41.78 & 4.46 & 35.10 & 83.44  & 86.64 & 90.06 & \textbf{68.61} & 58.81 & 55.52  &63.17   &26.65   &37.49  \\ \midrule
        BUS \cite{choe2024open} (MIC)                        &86.85 &43.49   & 89.35  &\textbf{46.12} &4.39  &54.29  & \textbf{87.90}   &92.49    &\textbf{91.46}  & 61.23  &58.11   & 59.81   &64.62   &33.37   &44.01   \\
        Ours (MIC)                       &87.27 & 49.47  & \textbf{89.50}  & 42.93   & \textbf{7.66}  & \textbf{60.16}  & 86.70   & \textbf{94.09} & 89.68 & 63.32 & \textbf{59.95} & \textbf{72.80}   &\textbf{66.96}   &\textbf{59.74}   &\textbf{62.65}  \\ 
			\bottomrule
		\end{tabular}
    \vspace{-15pt}
	\end{center}
\end{table*}

\begin{figure*}[!t]
	\centering
	\includegraphics[width=0.99\linewidth]{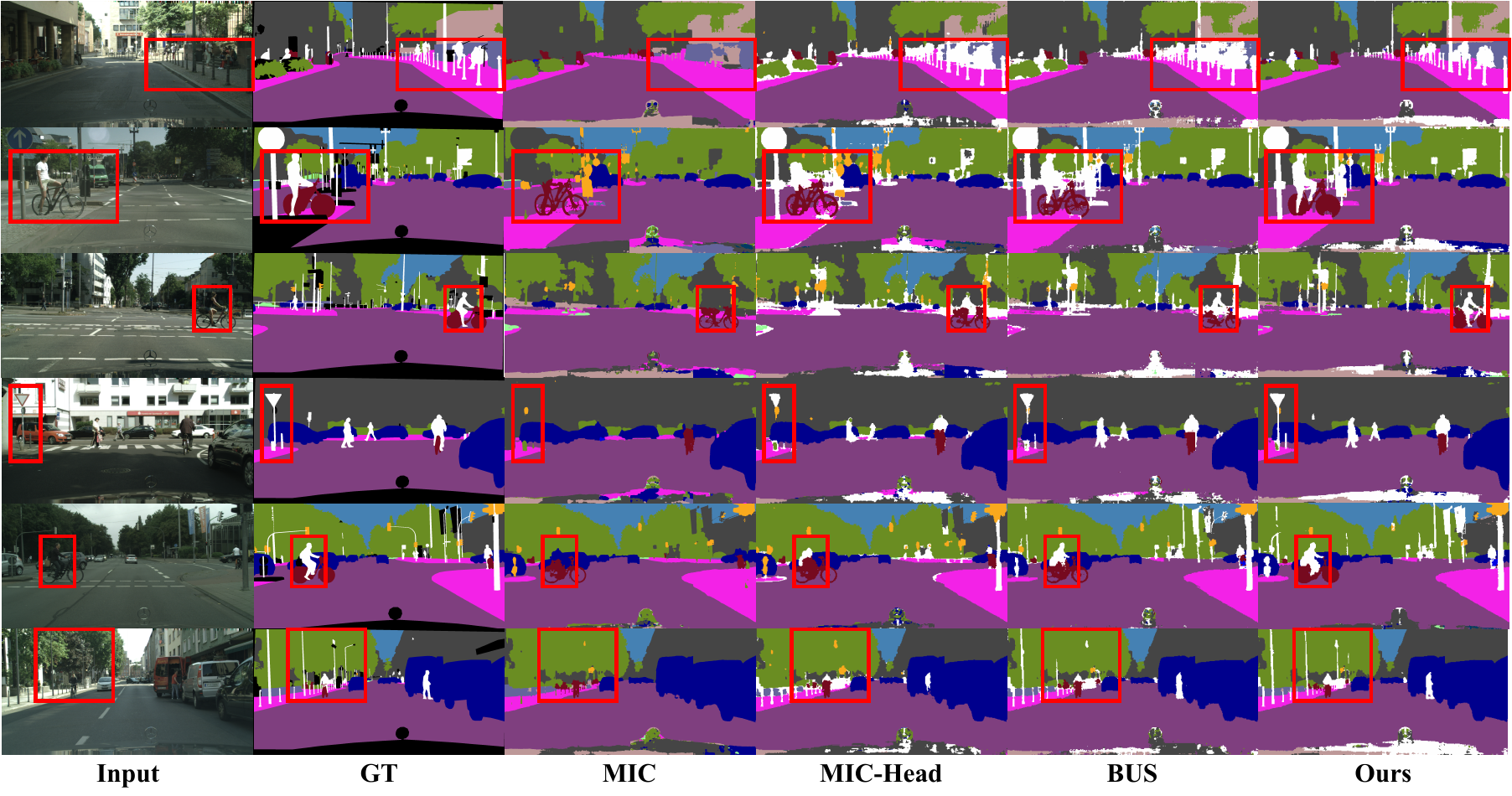}
	\caption{Visualization results of our method alongside competitive baselines, including the conventional CSDA-SS method MIC \cite{hoyer2023mic}, its head-expansion version (MIC-Head), and the OSDA-SS method BUS \cite{choe2024open}, on the GTA5$\rightarrow$Cityscapes benchmark. In these visualizations, white masks indicate unknown classes, and GT represents the ground truth.}
	\label{fig:resultsgta}
 \vspace{-5pt}
\end{figure*}

\begin{figure*}[!t]
  \centering
  \vspace{-5pt}
    \includegraphics[width=0.99\textwidth]{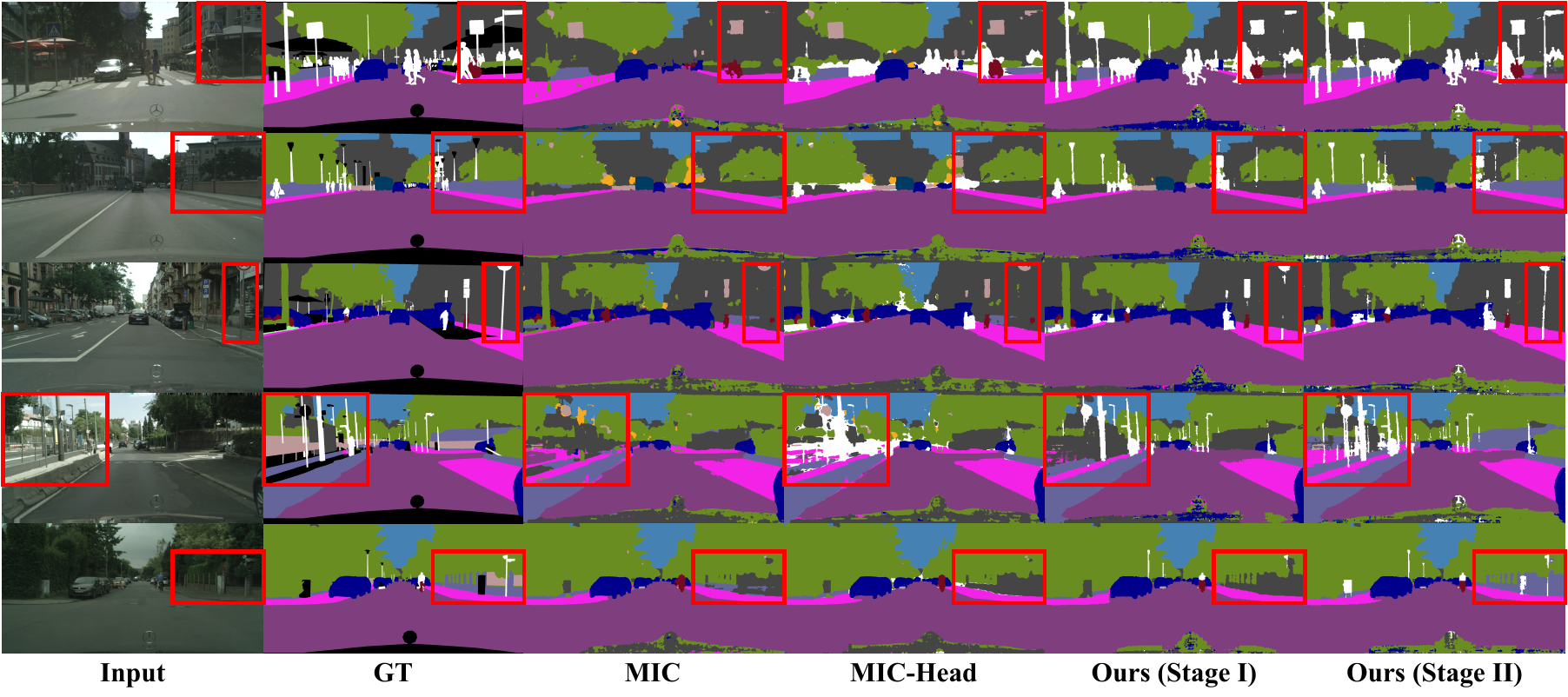}
    \caption{Qualitative comparison of our method between the first and second stages on the SYNTHIA $\rightarrow$ Cityscapes benchmark, alongside competitive baselines, including MIC \cite{hoyer2023mic} and its head-expansion version (MIC-Head).}
    \label{fig:synthia}
    \vspace{-5pt}
\end{figure*}

\begin{table}[!t]
	\begin{center}
		\fontsize{8}{11}\selectfont
		\caption{Ablation study of our proposed components on the GTA5 $\rightarrow$ Cityscapes benchmark.}
		\label{tab:ablation}
		\setlength{\tabcolsep}{4.5pt}
            \begin{tabular}{ccccccc}
			\hline
			\multirow{2}{*}{Config.} & \multicolumn{1}{c}{\multirow{2}{*}{\#Head}} & \multicolumn{1}{c}{Stage I} & \multicolumn{2}{c}{Stage II} & \multirow{2}{*}{Private} & \multirow{2}{*}{H-Score} \\ \cline{3-5}
			& \multicolumn{1}{c}{} & \multicolumn{1}{c}{VUC \quad } & ST \quad  & STH \quad & & \\ 
			\hline
			Config.A & K+1 &             &              &              & \cellcolor{gray!20}48.77         & \cellcolor{gray!20}57.88 \\
			Config.B & K+1 & $\checkmark$ &              &              & \cellcolor{gray!20}53.04         & \cellcolor{gray!20}61.38 \\
			Config.C & K+1 & $\checkmark$ & $\checkmark$ &              & \cellcolor{gray!20}57.74         & \cellcolor{gray!20}64.13 \\
			Config.D & K+1 & $\checkmark$ & $\checkmark$ & $\checkmark$ & \cellcolor{gray!20}\textbf{60.93} & \cellcolor{gray!20}\textbf{66.66} \\
			\hline
		\end{tabular}
	\end{center}
	\vspace{-8pt}
\end{table}

\begin{table}[!t]
	\begin{center}
		\fontsize{8}{11}\selectfont
		\caption{Performance improvements in existing methods with our SATS on the GTA5 $\rightarrow$ Cityscapes benchmark.}
		\label{tab:stage2}
		\setlength{\tabcolsep}{9pt}
		\begin{tabular}{ccccccc}
			\hline
            \multirow{2}{*}{Method}     &\multicolumn{2}{c}{BUS (NPL + MobileSAM)} & \multicolumn{2}{c}{Ours (VUC + MobileSAM)}& \multicolumn{2}{c}{Ours (VUC + SAM)}\\ \cline{2-7}
                     & Stage I & Stage II & Stage  I & Stage II  & Stage  I & Stage II \\ \hline
                    Private & \cellcolor{gray!20}50.30 & \cellcolor{gray!20}55.41 & \cellcolor{gray!20}54.39& \cellcolor{gray!20}58.40& \cellcolor{gray!20}53.04 & \cellcolor{gray!20}\textbf{60.93} \\
                    H-Score & \cellcolor{gray!20}58.83 & \cellcolor{gray!20}62.35 & \cellcolor{gray!20}61.05 & \cellcolor{gray!20}64.21 & \cellcolor{gray!20}61.38 & \cellcolor{gray!20}\textbf{66.66} \\
			\hline
		\end{tabular}
	\end{center}
	\vspace{-8pt}
\end{table}

\section{Experiments}
\subsection{Implementation}
\textbf{Training.} Our framework is based on the DAFormer architecture \cite{hoyer2022daformer}, equipped with the MiT-B5 encoder \cite{xie2021segformer}. We adopt the multi-resolution self-training strategy and training settings from MIC \cite{hoyer2023mic}. AdamW \cite{loshchilovdecoupled} serves as the optimizer, with learning rates set to 6e-5 for the backbone and 6e-4 for the decoder head. A weight decay of 0.01 is applied, and the learning rate is linearly warmed up over the first 1.5k steps. The predefined thresholds $\tau_1$ and $\tau_2$ are set to 0.5 and 0.968, respectively, and the smoothing factor $\alpha$ is set to 0.999. $\gamma$ is set to 0.25. We use SAM \cite{kirillov2023segment} to refine pseudo labels, following the refinement process outlined in \cite{choe2024open}. We incorporate ImageNet Feature Distance \cite{hoyer2022daformer}, Rare Class Sampling \cite{hoyer2022daformer}, DACS data augmentation \cite{tranheden2021dacs}, the Masked Image Consistency module \cite{hoyer2023mic}, and Thing Class Augmentation \cite{choe2024open}. Training for both stages runs over 40k iterations with a batch size of 2, using 512 × 512 random crops. The pre-training process takes 2k steps. 

\noindent\textbf{Benchmark construction.} To evaluate our framework on the OSDA-SS scenarios, we establish two synthetic-to-real benchmarks using existing self-driving datasets: GTA5 $\rightarrow$ Cityscapes and SYNTHIA $\rightarrow$ Cityscapes. The synthetic datasets include the GTA5 dataset \cite{richter2016playing}, which consists of 24,966 images, and the SYNTHIA dataset \cite{ros2016synthia}, with 9,400 images. The real-world dataset, Cityscapes \cite{cordts2016cityscapes}, contains 2,975 training samples and 500 validation samples. To introduce private classes unique to the target domain, we exclude specific classes from the source domain and reassign them to the ``ignore'' label to prevent their impact on training. The classes removed from the GTA5 dataset are ``pole'', ``traffic sign'', ``person'', ``rider'', ``truck'', and ``train''. In the SYNTHIA dataset, the excluded classes are ``pole'', ``traffic sign'', ``person'', ``rider'', ``truck'', ``train'', and ``terrain''. For evaluation purposes, these excluded classes are grouped as a single ``unknown'' class in Cityscapes.

\noindent\textbf{Metrics and baseline.} Following BUS \cite{choe2024open}, we employ three evaluation metrics: 1) the mean IoU score for known (common) classes, 2) the IoU score for the single unknown (private) class, and 3) the harmonic mean of the common mean IoU and private IoU scores, referred to as the H-Score. For the baselines, we first extend several classification methods originally designed for OSDA and universal domain adaptation to segmentation tasks, including OSBP \cite{saito2018open}, UAN \cite{you2019universal}, and UniOT \cite{jang2022unknown}. This extension is achieved by replacing the classification network with the DeepLabv2 architecture \cite{chen2017deeplab}, using ResNet-101 \cite{he2016deep} as the backbone. Second, we adapt CSDA segmentation methods--ASN \cite{tsai2018learning}, Pixmatch \cite{melas2021pixmatch}, DAF \cite{hoyer2022daformer}, HRDA \cite{hoyer2022hrda} and MIC \cite{hoyer2023mic}--for OSDA-SS by: 1) treating low-confidence pixels as ``unknown'' during inference based on a predefined threshold (confidence-threshold baseline), and 2) extending the classifier head from K to (K+1) classes during training (head-expansion baseline). Finally, we also compared our model with the existing OSDA-SS method, BUS \cite{choe2024open}.

\subsection{Comparison with state-of-the-art methods}
\label{sec:main_results}
We first evaluate our method against SOTA approaches. Table \ref{tab:STOA0} presents a detailed comparison with baseline methods on both OSDA-SS benchmarks. The results highlight limitations in current OSDA-SS strategies. Classification-based methods \cite{saito2018open,you2019universal,jang2022unknown}, when adapted for segmentation, often misclassify due to limited spatial awareness. Although CSDA-SS methods \cite{tsai2018learning,melas2021pixmatch,hoyer2022daformer,hoyer2022hrda,hoyer2023mic} provide a more effective solution than classification-based approaches, they still present significant limitations. In contrast, our proposed method exhibits superior performance over existing approaches. Notably, it surpasses the current SOTA OSDA-SS method \cite{choe2024open}, delivering significant performance improvements of +3.85\% in H-Score on GTA5 $\rightarrow$ Cityscapes and +18.64\% on SYNTHIA $\rightarrow$ Cityscapes. Figures \ref{fig:resultsgta} and \ref{fig:synthia} further supports these findings. It can be observed that existing methods often suffer from negative transfer, producing erroneous predictions for difficult/ambiguous known classes. As shown in the first row of Figure \ref{fig:resultsgta}, the model misclassifies the ``wall'' class as unknown. Additionally, due to annotation imbalance, the model learns discriminative features for known classes faster than for unknowns, causing some unknown classes to be misclassified as known classes. As illustrated in the third row of Figure \ref{fig:resultsgta}, the model erroneously predicts the ``pole'' class as ``car''. Instead, our proposed method addresses these problems, consistently delivering cleaner and more distinct segmentation masks, highlighting its robust and reliable performance in OSDA-SS scenarios. 

\begin{table}[t]
	\begin{center}
		\fontsize{8}{11}\selectfont
		\caption{The results with different numbers of target-private classes on the GTA5 $\rightarrow$ Cityscapes benchmark.}
		\label{tab:number}
		\setlength{\tabcolsep}{10pt}
		\begin{tabular}{cccc}
			\hline
            \multirow{2}{*}{\# of Novel} & \multirow{2}{*}{BUS \cite{choe2024open}} & \multicolumn{2}{c}{Ours} \\ \cline{3-4}
                                        & & Config. B & Config. D \\ \hline
            2               & \cellcolor{gray!20}56.82 & \cellcolor{gray!20}56.36 & \cellcolor{gray!20}\textbf{73.72} \\
            4               & \cellcolor{gray!20}54.72 & \cellcolor{gray!20}66.76 & \cellcolor{gray!20}\textbf{71.00} \\
            6               & \cellcolor{gray!20}62.81 & \cellcolor{gray!20}61.38 & \cellcolor{gray!20}\textbf{66.66} \\
            8               & \cellcolor{gray!20}62.01 & \cellcolor{gray!20}62.56 & \cellcolor{gray!20}\textbf{69.98} \\
            10              & \cellcolor{gray!20}55.56 & \cellcolor{gray!20}56.15 & \cellcolor{gray!20}\textbf{61.68} \\
			\hline
		\end{tabular}
	\end{center}
	\vspace{-8pt}
\end{table}

\begin{table}[t]
	\begin{center}
		\fontsize{8}{11}\selectfont
		\caption{Experiment results with different $C_H$ settings on the GTA5 $\rightarrow$ Cityscapes benchmark.}
		\label{tab:c_h}
		\setlength{\tabcolsep}{9pt}
		\begin{tabular}{cccc}
			\hline
            $C_H$                 & Common & Private & H-Score  \\ \hline
            $C_S$                & \cellcolor{gray!20}73.00   & \cellcolor{gray!20}59.32   & \cellcolor{gray!20}65.45 \\
            $C_S$-Tail classes   & \cellcolor{gray!20}72.27   & \cellcolor{gray!20}57.90   & \cellcolor{gray!20}64.29 \\
            $C_S$-Head classes   & \cellcolor{gray!20}\textbf{73.57}   & \cellcolor{gray!20}\textbf{60.93}   & \cellcolor{gray!20}\textbf{66.66} \\
			\hline
		\end{tabular}
	\end{center}
	\vspace{-8pt}
\end{table}

\begin{table}[t]
	\begin{center}
		\fontsize{8}{11}\selectfont
		\caption{Quantitative comparison with randomly selected private classes on the GTA5$\rightarrow$Cityscapes benchmark. We performed three experiments and reported the average deviation.}
		\label{tab:random}
		\setlength{\tabcolsep}{9pt}
		\begin{tabular}{cccc}
			\hline
            Method                & MIC \cite{hoyer2023mic}     & BUS \cite{choe2024open}    & Ours  \\ \hline
            H-Score               & \cellcolor{gray!20}46.24$\pm$4.94   & \cellcolor{gray!20}53.75$\pm$14.31   & \cellcolor{gray!20}\textbf{61.63$\pm$5.97} \\
			\hline
		\end{tabular}
	\end{center}
	\vspace{-8pt}
\end{table}

\begin{table}[h!]
	\begin{center}
		\fontsize{8}{11}\selectfont
		\caption{Sensitivity analysis of $\tau_1$ on the GTA5 $\rightarrow$ Cityscapes benchmark.}
		\label{tab:sensitivity}
		\setlength{\tabcolsep}{7pt}
		\begin{tabular}{cccccc}
			\hline
            $\tau_1$  &0.3   &0.4   & 0.5   & 0.6  & 0.7  \\ \hline
            Common    &\cellcolor{gray!20} 73.08     &\cellcolor{gray!20} 71.96     & \cellcolor{gray!20}\textbf{73.57} &\cellcolor{gray!20} 62.23     &\cellcolor{gray!20} 53.97     \\
            Private   &\cellcolor{gray!20} 52.86     &\cellcolor{gray!20} 58.43     & \cellcolor{gray!20}\textbf{63.93} &\cellcolor{gray!20} 28.55     &\cellcolor{gray!20}19.72      \\
            H-Score   &\cellcolor{gray!20} 61.35     &\cellcolor{gray!20} 64.49     & \cellcolor{gray!20}\textbf{66.66} &\cellcolor{gray!20} 39.15     &\cellcolor{gray!20} 28.89     \\
			\hline
		\end{tabular}
	\end{center}
	\vspace{-8pt}
\end{table}

\subsection{Ablation study}
\textbf{Component-wise ablation.} In this section, we begin with ablation experiments to validate the effectiveness of the proposed components. The results, shown in Table \ref{tab:ablation}, offer a detailed breakdown. In this table, ``\#Head'' refers to the dimensionality of the classifier head, while ``VUC'' and ``ST'' represent virtual unknown construction and self-training without hard unknown exploration, respectively. The complete unknown-aware domain adaptation process is denoted as ``STH'', where hard unknown exploration is incorporated into the self-training process. 
The results show that the complete implementation of the proposed method achieves state-of-the-art performance (config.D).
By creating virtual unknowns to facilitate training of the expanded head (config.B), we increase the H-Score from 57.88\% to 61.38\%. Moreover, the additional performance gains achieved through ST--where we introduce target unknowns initially predicted in Stage I into the source domain (config.C)--strongly support our illustration that imbalanced annotations between known and unknown classes lead to negative transfer of known classes and underfitting of unknowns. This step effectively mitigates these limitations and further boosts the model's performance. In addition, by dynamically exploring hard unknowns and augmenting them within source samples (config.D), our method yields even greater improvements, raising the H-Score from 64.13\% to 66.66\%.

\noindent\textbf{Noisy pseudo labels vs. virtual unknowns.} In BUS \cite{choe2024open}, noisy pseudo labels (NPL) are utilized to facilitate training of the expanded head. In contrast, we tackle this issue by constructing virtual unknowns (VUC).  To validate the advantages of our VUC, we conduct additional experiments in this section. To ensure a fair comparison, we remove the DECON loss from BUS. For our proposed method, we replace the refinement model from SAM \cite{kirillov2023segment} with MobileSAM \cite{zhang2023faster}. The results are displayed in Table \ref{tab:stage2} (column 2 vs. column 4). Our findings indicate that the proposed VUC is more effective in training the expanded head, yielding superior outcomes in the detection of private classes and enhancing overall performance.

\noindent\textbf{Influence of our SATS with different methods.} Table \ref{tab:stage2} also presents the performance gains achieved in existing methods with the application of our SATS. It can be observed that by further applying our unknown-aware domain adaptation method, existing one-stage baselines achieve consistent performance improvements (column 2 vs. column 3). Moreover, our method consistently outperforms BUS in both stages(column 2 vs. column 4; column 3 vs. column 5), suggesting that our approach is more robust and adaptable in handling unknown classes.

\subsection{Sensitivity analysis of parameters}

\noindent\textbf{Proportion of unknown classes.} We further evaluate the impact of different numbers of target unknowns on model performance. The experimental results, summarized in Table \ref{tab:number}, provide an overview of this effect. The selection of target-private classes under different conditions follows the protocol established in BUS \cite{choe2024open}. Our experiments show that, regardless of whether the number of unknown classes is increased or decreased, our method consistently achieves notable performance improvements compared to other approaches. This trend underscores the robustness of our approach across diverse scenarios and highlights its adaptability to varying task complexities.

\noindent\textbf{Impact of $C_H$.} We assess the influence of different choices for $C_H$. As demonstrated in Table \ref{tab:c_h}, we configure $C_H$ to encompass all known classes ($C_S$), the tail classes within the known classes ($C_S$-Tail classes), and the head classes within the known classes ($C_S$-Head classes). The head and tail classes are defined by class frequencies as DAFormer \cite{hoyer2022daformer}.  Head classes, with higher frequencies, are ``road'', ``sidewalk'', ``building'', ``vegetation'', ``sky'', and ``car'', while the remaining classes in $C_S$ are tail classes. The results indicate that top performance is achieved when $C_H$ is set to $C_S$-Head classes.

\noindent\textbf{Selection of $C_{T\setminus S}$.} In the main experiments, thing classes are selected as the unknown classes $C_{T\setminus S}$. To further investigate the impact of $C_{T\setminus S}$ on model performance, we also include stuff classes in the selected private classes. Specifically, 6 classes are randomly chosen from the 19 available classes, regardless of whether they are thing or stuff categories. For a fair comparison, we retrain MIC \cite{hoyer2023mic} and BUS \cite{choe2024open} under the same conditions. The results shown in Table \ref{tab:random}  confirm the superiority of our method, demonstrating its robustness across different class compositions.

\noindent\textbf{Influence of $\tau_1$.} In the head-expansion baseline, we use $\tau_1$ as the predefined threshold to reassign low-confidence pixels as ``unknown''. In this section, we analyze the impact of varying $\tau_1$ on the overall performance. The experimental results are summarized in Table \ref{tab:sensitivity}, where we observe that the model achieves its best performance when $\tau_1$ is set to 0.5. This suggests that a balanced threshold value effectively distinguishes low-confidence pixels as ``unknown'' without overly penalizing the segmentation of known classes. Higher values of $\tau_1$ may result in excessive misclassification of known pixels as unknown, while lower values may fail to adequately capture true unknown pixels, leading to suboptimal performance. Thus, $\tau_1 = 0.5$ strikes a favorable trade-off between these factors, highlighting its importance in achieving optimal model performance.

\section{Conclusion}
In this paper, we propose SATS, a Separating-then-Adapting Training Strategy designed to address OSDA-SS through two sequential steps: known/unknown separation and unknown-aware domain adaptation. Additionally, we propose hard unknown exploration, a new data augmentation method that exposes the model to more challenging unknowns, thereby enhancing its ability to learn more distinct features. We assess the performance of our method on the public OSDA-SS benchmarks, demonstrating that it significantly surpasses other competing methods. We anticipate that our approach to improve safety and reliability in dynamic environments like autonomous driving, healthcare, and robotics, laying a foundation for future AI advancements in unknown detection. 

\section{Limitations and Future Works}
\label{limi}
\textbf{Limitations.} While our proposed method demonstrates strong performance across various benchmarks, there are certain limitations that warrant further investigation. Specifically, our current approach focuses on a simplified scenario where the source and target samples are assumed to originate from static distributions. However, real-world systems are often dynamic, with continuous and unpredictable distribution shifts occurring over time. This limitation constrains the method’s applicability in scenarios where the data evolves, such as autonomous driving in changing weather conditions or adaptive systems responding to user behavior. 

\noindent\textbf{Future works.} To address this limitation, our future work will focus on developing robust methodologies to tackle OSDA-SS under continuous distribution shifts. This involves designing mechanisms to adaptively update the model in response to distributional changes, ensuring its ability to generalize effectively across evolving environments. Additionally, integrating techniques for detecting and handling novel unknown classes that emerge during such shifts will be a key area of focus. By addressing these challenges, we aim to extend the applicability of our method to more dynamic and realistic scenarios.







\bibliography{main}

@String(AAAI = {AAAI})

@article{goodfellow2014generative,
  title={Generative adversarial nets},
  author={Goodfellow I and Pouget-Abadie J and Mirza M and others},
  journal={Advances in neural information processing systems},
  volume={27},
  year={2014}
}

@inproceedings{hoffman2018cycada,
  title={Cy{CADA}: Cycle-consistent adversarial domain adaptation},
  author={Hoffman J and Tzeng E and Park T and others},
  booktitle={Proceedings of the International Conference on Machine Learning},
  pages={1989--1998},
  year={2018},
  organization={PMLR}
}

@inproceedings{pizzati2020domain,
  title={Domain bridge for unpaired image-to-image translation and unsupervised domain adaptation},
  author={Pizzati F and Charette R and Zaccaria M and others},
  booktitle={Proceedings of the IEEE/CVF Winter Conference on Applications of Computer Vision},
  pages={2990--2998},
  year={2020}
}

@article{gong2021dlow,
  title={{DLOW}: Domain flow and applications},
  author={Gong R and Li W and Chen Y and others},
  journal={International Journal of Computer Vision},
  volume={129},
  number={10},
  pages={2865--2888},
  year={2021},
  publisher={Springer}
}

@article{ganin2016domain,
	title={Domain-adversarial training of neural networks},
	author={Ganin Y and Ustinova E and Ajakan H and others},
	journal={The Journal of Machine Learning Research},
	volume={17},
	number={1},
	pages={2096--2030},
	year={2016}
}

@article{long2018conditional,
	title={Conditional adversarial domain adaptation},
	author={Long M and Cao Z and Wang J and others},
	journal={Advances in Neural Information Processing Systems},
	volume={31},
	year={2018}
}

@article{zhang2023faster,
  title={Faster segment anything: Towards lightweight sam for mobile applications},
  author={Zhang C and Han D and Qiao Y and others},
  journal={arXiv preprint arXiv:2306.14289},
  year={2023}
}

@inproceedings{tsai2018learning,
	title={Learning to adapt structured output space for semantic segmentation},
	author={Tsai Y-H and Hung W-C and Schulter S and others},
	booktitle={Proceedings of the IEEE Conference on Computer Vision and Pattern Recognition},
	pages={7472--7481},
	year={2018}
}

@inproceedings{saito2018maximum,
	title={Maximum classifier discrepancy for unsupervised domain adaptation},
	author={Saito K and Watanabe K and Ushiku Y and others},
	booktitle={Proceedings of the IEEE Conference on Computer Vision and Pattern Recognition},
	pages={3723--3732},
	year={2018}
}

@inproceedings{vu2019advent,
	title={{ADVENT}: Adversarial entropy minimization for domain adaptation in semantic segmentation},
	author={Vu T-H and Jain H and Bucher M and others},
	booktitle={Proceedings of the IEEE/CVF Conference on Computer Vision and Pattern Recognition},
	pages={2517--2526},
	year={2019}
}

@article{luo2021category,
	title={Category-level adversarial adaptation for semantic segmentation using purified features},
	author={Luo Y and Liu P and Zheng L and others},
	journal={IEEE Transactions on Pattern Analysis and Machine Intelligence},
	volume={44},
	number={8},
	pages={3940--3956},
	year={2021},
	publisher={IEEE}
}

@inproceedings{mei2020instance,
	title={Instance adaptive self-training for unsupervised domain adaptation},
	author={Mei K and Zhu C and Zou J and others},
	booktitle={Proceedings of the European Conference on Computer Vision},
	pages={415--430},
	year={2020},
	organization={Springer}
}

@inproceedings{zhang2018collaborative,
	title={Collaborative and adversarial network for unsupervised domain adaptation},
	author={Zhang W and Ouyang W and Li W and others},
	booktitle={Proceedings of the IEEE Conference on Computer Vision and Pattern Recognition},
	pages={3801--3809},
	year={2018}
}

@inproceedings{zou2018unsupervised,
	title={Unsupervised domain adaptation for semantic segmentation via class-balanced self-training},
	author={Zou Y and Yu Z and Vijaya Kumar B V K  and others},
	booktitle={Proceedings of the European Conference on Computer Vision},
	pages={289--305},
	year={2018}
}

@inproceedings{zhang2021prototypical,
  title={Prototypical pseudo label denoising and target structure learning for domain adaptive semantic segmentation},
  author={Zhang P and Zhang B and Zhang T and others},
  booktitle={Proceedings of the IEEE/CVF Conference on Computer Vision and Pattern Recognition},
  pages={12414--12424},
  year={2021}
}

@article{zhang2019category,
  title={Category anchor-guided unsupervised domain adaptation for semantic segmentation},
  author={Zhang Q and Zhang J and Liu W and others},
  journal={Advances in Neural Information Processing Systems},
  volume={32},
  year={2019}
}

@inproceedings{araslanov2021self,
	title={Self-supervised augmentation consistency for adapting semantic segmentation},
	author={Araslanov N and Roth S},
	booktitle={Proceedings of the IEEE/CVF Conference on Computer Vision and Pattern Recognition},
	pages={15384--15394},
	year={2021}
}

@inproceedings{choi2019self,
	title={Self-ensembling with {GAN}-based data augmentation for domain adaptation in semantic segmentation},
	author={Choi J and Kim T and Kim C},
	booktitle={Proceedings of the IEEE/CVF International Conference on Computer Vision},
	pages={6830--6840},
	year={2019}
}

@article{zhou2022context,
	title={Context-aware mixup for domain adaptive semantic segmentation},
	author={Zhou Q and Feng Z and Gu Q and others},
	journal={IEEE Transactions on Circuits and Systems for Video Technology},
	volume={33},
	number={2},
	pages={804--817},
	year={2022},
	publisher={IEEE}
}

@inproceedings{hoyer2022daformer,
	title={{DAF}ormer: Improving network architectures and training strategies for domain-adaptive semantic segmentation},
	author={Hoyer L and Dai D and Van Gool L},
	booktitle={Proceedings of the IEEE/CVF Conference on Computer Vision and Pattern Recognition},
	pages={9924--9935},
	year={2022}
}

@inproceedings{kim2023bidirectional,
	title={Bidirectional Domain Mixup for Domain Adaptive Semantic Segmentation},
	author={Kim D and Seo M and Park K and others},
	booktitle={Proceedings of the AAAI Conference on Artificial Intelligence},
	volume={37},
	pages={1114--1123},
	year={2023}
}

@inproceedings{hoyer2022hrda,
	title={{HRDA}: Context-aware high-resolution domain-adaptive semantic segmentation},
	author={Hoyer L and Dai D and Van Gool L},
	booktitle={Proceedings of the European Conference on Computer Vision},
	pages={372--391},
	year={2022}
}

@article{zhang2021multiple,
  title={Multiple fusion adaptation: A strong framework for unsupervised semantic segmentation adaptation},
  author={Zhang K and Sun Y and Wang R and others},
  journal={arXiv preprint arXiv:2112.00295},
  year={2021}
}

@article{zhou2022uncertainty,
  title={Uncertainty-aware consistency regularization for cross-domain semantic segmentation},
  author={Zhou Q and Feng Z and Gu Q and others},
  journal={Computer Vision and Image Understanding},
  volume={221},
  pages={103448},
  year={2022},
  publisher={Elsevier}
}

@article{zheng2021rectifying,
  title={Rectifying pseudo label learning via uncertainty estimation for domain adaptive semantic segmentation},
  author={Zheng Z and Yang Y},
  journal={International Journal of Computer Vision},
  volume={129},
  number={4},
  pages={1106--1120},
  year={2021},
  publisher={Springer}
}

@inproceedings{kim2020learning,
	title={Learning texture invariant representation for domain adaptation of semantic segmentation},
	author={Kim M and Byun H},
	booktitle={Proceedings of the IEEE/CVF Conference on Computer Vision and Pattern Recognition},
	pages={12975--12984},
	year={2020}
}

@inproceedings{wang2020classes,
  title={Classes matter: A fine-grained adversarial approach to cross-domain semantic segmentation},
  author={Wang H and Shen T and Zhang W and others},
  booktitle={Proceedings of the European Conference on Computer Vision},
  pages={642--659},
  year={2020},
  organization={Springer}
}

@inproceedings{zheng2021unsupervised,
  title={Unsupervised scene adaptation with memory regularization in vivo},
  author={Zheng Z and Yang Y},
  booktitle={Proceedings of the International Joint Conference on Artificial Intelligence},
  year={2021}
}

@inproceedings{liu2021bapa,
  title={{BAPA-N}net: Boundary adaptation and prototype alignment for cross-domain semantic segmentation},
  author={Liu Y and Deng J and Gao X and others},
  booktitle={Proceedings of the IEEE/CVF International Conference on Computer Vision},
  pages={8801--8811},
  year={2021}
}

@inproceedings{huang2022category,
	title={Category contrast for unsupervised domain adaptation in visual tasks},
	author={Huang J and Guan D and Xiao A and others},
	booktitle={Proceedings of the IEEE/CVF Conference on Computer Vision and Pattern Recognition},
	pages={1203--1214},
	year={2022}
}

@article{xie2023sepico,
	title={Se{PICO}: Semantic-guided pixel contrast for domain adaptive semantic segmentation},
	author={Xie B and Li S and Li M and others},
	journal={IEEE Transactions on Pattern Analysis and Machine Intelligence},
	year={2023},
	publisher={IEEE}
}

@inproceedings{kundu2020towards,
  title={Towards inheritable models for open-set domain adaptation},
  author={Kundu J N and Venkat N and Revanur A and others},
  booktitle={Proceedings of the IEEE/CVF Conference on Computer Vision and Pattern Recognition},
  pages={12376--12385},
  year={2020}
}

@inproceedings{tsai2019domain,
	title={Domain adaptation for structured output via discriminative patch representations},
	author={Tsai Y-H and Sohn K and Schulter S and others},
	booktitle={Proceedings of the IEEE/CVF International Conference on Computer Vision},
	pages={1456--1465},
	year={2019}
}

@inproceedings{panareda2017open,
  title={Open set domain adaptation},
  author={Panareda Busto P and Gall J},
  booktitle={Proceedings of the IEEE International Conference on Computer Vision},
  pages={754--763},
  year={2017}
}

@inproceedings{luo2020progressive,
  title={Progressive graph learning for open-set domain adaptation},
  author={Luo Y and Wang Z and Huang Z and others},
  booktitle={Proceedings of the International Conference on Machine Learning},
  pages={6468--6478},
  year={2020},
  organization={PMLR}
}

@inproceedings{liu2019separate,
  title={Separate to adapt: Open set domain adaptation via progressive separation},
  author={Liu H and Cao Z and Long M and others},
  booktitle={Proceedings of the IEEE/CVF Conference on Computer Vision and Pattern Recognition},
  pages={2927--2936},
  year={2019}
}

@article{wang2024progressively,
  title={Progressively select and reject pseudo-labelled samples for open-set domain adaptation},
  author={Wang Q and Meng F and Breckon T P},
  journal={IEEE Transactions on Artificial Intelligence},
  year={2024},
  publisher={IEEE}
}

@inproceedings{choe2024open,
  title={Open-Set Domain Adaptation for Semantic Segmentation},
  author={Cho S-A and Shin A-H and Park K-H and others},
  booktitle={Proceedings of the IEEE/CVF Conference on Computer Vision and Pattern Recognition},
  pages={23943--23953},
  year={2024}
}

@inproceedings{saito2018open,
  title={Open set domain adaptation by backpropagation},
  author={Saito K and Yamamoto S and Ushiku Y and others},
  booktitle={Proceedings of the European Conference on Computer Vision},
  pages={153--168},
  year={2018}
}

@inproceedings{you2019universal,
  title={Universal domain adaptation},
  author={You K and Long M and Cao Z and others},
  booktitle={Proceedings of the IEEE/CVF Conference on Computer Vision and Pattern Recognition},
  pages={2720--2729},
  year={2019}
}

@article{jang2022unknown,
  title={Unknown-aware domain adversarial learning for open-set domain adaptation},
  author={Jang J and Na B and Shin D H and others},
  journal={Advances in Neural Information Processing Systems},
  volume={35},
  pages={16755--16767},
  year={2022}
}

@inproceedings{hoyer2023mic,
  title={{MIC}: Masked image consistency for context-enhanced domain adaptation},
  author={Hoyer L and Dai D  and Wang H and others},
  booktitle={Proceedings of the IEEE/CVF Conference on Computer Vision and Pattern Recognition},
  pages={11721--11732},
  year={2023}
}

@article{xie2021segformer,
  title={{S}eg{F}ormer: Simple and efficient design for semantic segmentation with transformers},
  author={Xie E and Wang W and Yu Z and others},
  journal={Advances in Neural Information Processing Systems},
  volume={34},
  pages={12077--12090},
  year={2021}
}

@inproceedings{loshchilovdecoupled,
  title={Decoupled Weight Decay Regularization},
  author={Loshchilov I and Hutter F},
  booktitle={Proceedings of the International Conference on Learning Representations},
  year={2018}
}

@inproceedings{tranheden2021dacs,
  title={{DACS}: Domain adaptation via cross-domain mixed sampling},
  author={Tranheden W and Olsson V and Pinto J and others},
  booktitle={Proceedings of the IEEE/CVF Winter Conference on Applications of Computer Vision},
  pages={1379--1389},
  year={2021}
}

@inproceedings{kirillov2023segment,
  title={Segment anything},
  author={Kirillov A and Mintun E and Ravi N and others},
  booktitle={Proceedings of the IEEE/CVF International Conference on Computer Vision},
  pages={4015--4026},
  year={2023}
}

@inproceedings{richter2016playing,
  title={Playing for data: Ground truth from computer games},
  author={Richter S R and Vineet V and Roth S and others},
  booktitle={Proceedings of the European Conference on Computer Vision},
  pages={102--118},
  year={2016},
  organization={Springer}
}

@inproceedings{ros2016synthia,
  title={The {SYNTHIA} dataset: A large collection of synthetic images for semantic segmentation of urban scenes},
  author={Ros G and Sellart L and Materzynska J and others},
  booktitle={Proceedings of the IEEE Conference on Computer Vision and Pattern Recognition},
  pages={3234--3243},
  year={2016}
}

@inproceedings{cordts2016cityscapes,
  title={The cityscapes dataset for semantic urban scene understanding},
  author={Cordts M and Omran M and Ramos S and others},
  booktitle={Proceedings of the IEEE Conference on Computer Vision and Pattern Recognition},
  pages={3213--3223},
  year={2016}
}

@article{chen2017deeplab,
  title={Deep{L}ab: Semantic image segmentation with deep convolutional nets, atrous convolution, and fully connected {CRF}s},
  author={Chen L-C and Papandreou G and Kokkinos I and others},
  journal={IEEE Transactions on Pattern Analysis and Machine Intelligence},
  volume={40},
  number={4},
  pages={834--848},
  year={2017},
  publisher={IEEE}
}

@inproceedings{he2016deep,
  title={Deep residual learning for image recognition},
  author={He K and Zhang X and Ren S and others},
  booktitle={Proceedings of the IEEE Conference on Computer Vision and Pattern Recognition},
  pages={770--778},
  year={2016}
}

@inproceedings{sakaridis2021acdc,
  title={{ACDC}: The adverse conditions dataset with correspondences for semantic driving scene understanding},
  author={Sakaridis C and Dai D and Van Gool L},
  booktitle={Proceedings of the IEEE/CVF International Conference on Computer Vision},
  pages={10765--10775},
  year={2021}
}

@inproceedings{melas2021pixmatch,
  title={Pix{M}atch: Unsupervised domain adaptation via pixelwise consistency training},
  author={Melas-Kyriazi L and Manrai A K},
  booktitle={Proceedings of the IEEE/CVF Conference on Computer Vision and Pattern Recognition},
  pages={12435--12445},
  year={2021}
}

@inproceedings{liu2019auto,
  title={Auto-{D}eep{L}ab: Hierarchical neural architecture search for semantic image segmentation},
  author={Liu C and Chen L-C and Schroff F and others},
  booktitle={Proceedings of the IEEE/CVF Conference on Computer Vision and Pattern Recognition},
  pages={82--92},
  year={2019}
}

@inproceedings{strudel2021segmenter,
  title={Segmenter: Transformer for semantic segmentation},
  author={Strudel R and Garcia R and Laptev I and others},
  booktitle={Proceedings of the IEEE/CVF International Conference on Computer Vision},
  pages={7262--7272},
  year={2021}
}

@inproceedings{liu2020negative,
  title={Negative margin matters: Understanding margin in few-shot classification},
  author={Liu B and Cao Y and Lin Y and others},
  booktitle={Proceedings of the European Conference on Computer Vision},
  pages={438--455},
  year={2020},
  organization={Springer}
}

@inproceedings{
    cao2022openworld,
    title={Open-World Semi-Supervised Learning},
    author={Cao K  and Brbic M  and Leskovec J},
    booktitle={Proceedings of the International Conference on Learning Representations},
    year={2022},
    url={https://openreview.net/forum?id=O-r8LOR-CCA}
}

@article{saito2020universal,
  title={Universal domain adaptation through self supervision},
  author={Saito K and Kim D and Sclaroff S and others},
  journal={Advances in Neural Information Processing Systems},
  volume={33},
  pages={16282--16292},
  year={2020}
}
\bibliographystyle{unsrt}



\end{document}